\documentclass[letterpaper]{article} 
\usepackage[submission]{aaai24}  
\usepackage{times}  
\usepackage{helvet}  
\usepackage{courier}  
\usepackage[hyphens]{url}  
\usepackage{graphicx} 
\usepackage{natbib}  
\usepackage{caption} 
\usepackage{algorithm}
\usepackage{algorithmic}

\usepackage{newfloat}
\usepackage{listings}
\DeclareCaptionStyle{ruled}{labelfont=normalfont,labelsep=colon,strut=off} 
\floatstyle{ruled}
\newfloat{listing}{tb}{lst}{}
\floatname{listing}{Listing}
\title{Online Submission and Evaluation System Design for Competition Operations}
\author{
    Zhe Chen\textsuperscript{\rm 1},
    Daniel Harabor\textsuperscript{\rm 1},
    Ryan Hechnenberger\textsuperscript{\rm 1},
    Nathan R. Sturtevant\textsuperscript{\rm 2}
}
\affiliations{
    \textsuperscript{\rm 1} Department of Data Science and Artificial Intelligence, Monash University, Australia\\
    \textsuperscript{\rm 2}  Department of Computing Science, University of Alberta, Canada\\
}

\usepackage{bibentry}

\begin{document}

\maketitle

\begin{abstract}
Research communities have developed benchmark datasets across domains to compare the performance of algorithms and techniques
However, tracking the progress in these research areas is not easy,
as publications appear in different venues at the same time,
and many of them claim to represent the state-of-the-art.
To address this, research communities often organise periodic competitions to 
evaluate the performance of various algorithms and techniques, 
thereby tracking advancements in the field.
However, these competitions pose a significant operational burden. 
The organisers must manage and evaluate a large volume of submissions. 
Furthermore, participants typically develop their solutions in diverse environments, 
leading to compatibility issues during the evaluation of their submissions.
This paper presents an online competition system that automates the submission 
and evaluation process for a competition.
The competition system allows organisers to manage large numbers of submissions efficiently,
utilising isolated environments to evaluate submissions.
This system has already been used successfully for several competitions, including the Grid-Based Pathfinding Competition and the League of Robot Runners competition.
\end{abstract}

\section{Introduction}

Research on solving combinatorial optimisation problems has been popular
in the field of computer science and artificial intelligence.
Researchers from different domains have proposed various algorithms and techniques
to solve different types of problems.
To compare the performance of different algorithms and techniques,
research communities have developed several benchmark datasets, including
the 2D Path Finding Benchmarks~\cite{sturtevant2012benchmarks},
Iron Harvest Path Finding Benchmarks~\cite{iron},
Multi-Agent Path Finding Benchmarks~\cite{stern2019mapf},
Voxel Benchmarks for 3D Path Finding~\cite{brewer2018voxels,3d-pathfinding},
and so on.

However, tracking the progress in these research areas is not easy,
as publications appear in different venues at the same time,
and many of them claim to represent the state-of-the-art.
To address this issue, research communities have organised competitions
to evaluate the performance of different algorithms and techniques~\cite{gppc,ipc}.
These competitions often come with new benchmark problems and datasets
to boost the research in the field.
However, the operational burden of these competitions is high,
as the competition organisers need to manage a large number of submissions, conduct performance benchmarks across all the submissions, 
and provide feedback to the participants.
Moreover, participants often develop their solutions in different environments,
which leads to all sorts of compatibility issues when evaluating their submissions.

To address these issues, this paper presents an online competition system, which automates
the submission and evaluation process of a competition.
The system allows participants to submit their solutions to the competition problems at any time before the due date and evaluates the submissions in an isolated environment.
The instant feedback provided by the system allows participants to iterate their solutions quickly and efficiently. 
The system helped to bring back the Grid-based Path Planning Competition (GPPC$^2$)\footnote{https://gppc.search-conference.org/} in 2023, which was last held in 2014 and discontinued due to the high operation burden. 
The same architecture also serves the League of 
Robot Runners competition\footnote{https://www.leagueofrobotrunners.org/}, which attracts more than 25 active teams worldwide with a total of 825 submissions.
The competition system receives on average 9 submissions per day with a peak of 35 submissions 
per day during the competition period.

In this paper, we will first introduce the architecture of the competition system.
Then talk about how this system is used across different applications.
While running these competitions, we have faced several issues and challenges,
we will discuss and share our experiences in handling them.

\section{Related Works}
There are several existing approaches to competition hosting and benchmarking.
A number of these approaches are designed for specific types of problems or serving specific purposes:
\begin{itemize}
\item StarExec\footnote{https://github.com/StarExec/StarExec} is an open-source solving service developed at the University of Iowa, which allows benchmarking and comparing solvers on customised benchmark problems. 
However, the main goal of this project is to facilitate the experimental evaluation of solvers, which lack functionalities for competition management.
Competitions using StarExec manually process the evaluation results downloaded from the platform for ranking and results announcement.

\item DOMjudge\footnote{https://github.com/DOMjudge/domjudge} is an open-source programming contest jury system for International Collegiate Programming Contest (ICPC) style competitions. 
Although DOMjudge comes with a competition management system, the system is designed to evaluate simple algorithm implementations, usually a program in one file, in a restricted programming environment. 
The system requires competition problems documented in certain formats and judges submitted programs mainly in output correctness.

\item CodaLab\footnote{https://codalab.lisn.upsaclay.fr} and CodaBench\footnote{https://www.codabench.org} are open-source competition platforms that allow competition organisers to facilitate competitions on the platform or deploy the platform on their own.
The platform allows competition organisers to describe their competition using a competition ``bundle'', which includes a YAML file describing the metadata of the competition, HTML pages for displaying competition descriptions, an ingestion program that executes participant codes, a scoring program that evaluates and scores the output of participant codes, and other supporting files.
Although the platform can host any competition that can be described within the rules of the competition ``bundle'', but customising the platform for non-standard competition workflow (e.g. the GPPC parallels precomputing but only allows one benchmarking process per evaluation worker) is hard due to the complex platform implementation.
While one aim of this work is to provide the community a light weight system that is easy for customisation.

\end{itemize}

Our software aims to host competitions where solvers output complex solutions, and each submission may consist of many files and can use any dependent libraries. 
The performance metrics we measure are more than correctness, which includes, but are not limited to, solution quality, runtime, storage usage, RAM usage, pre-processing time, and so on. Different competitions we host measure submissions differently and require different evaluation processes.
Modifying StartExec or DOMjudge to run our competitions is not trivial, we thus designed our competition system and aimed to provide the research community with an open-source and customisable solution.

\section{System Architecture}

\begin{figure}[t]
\centering
\includegraphics[width=1\columnwidth]{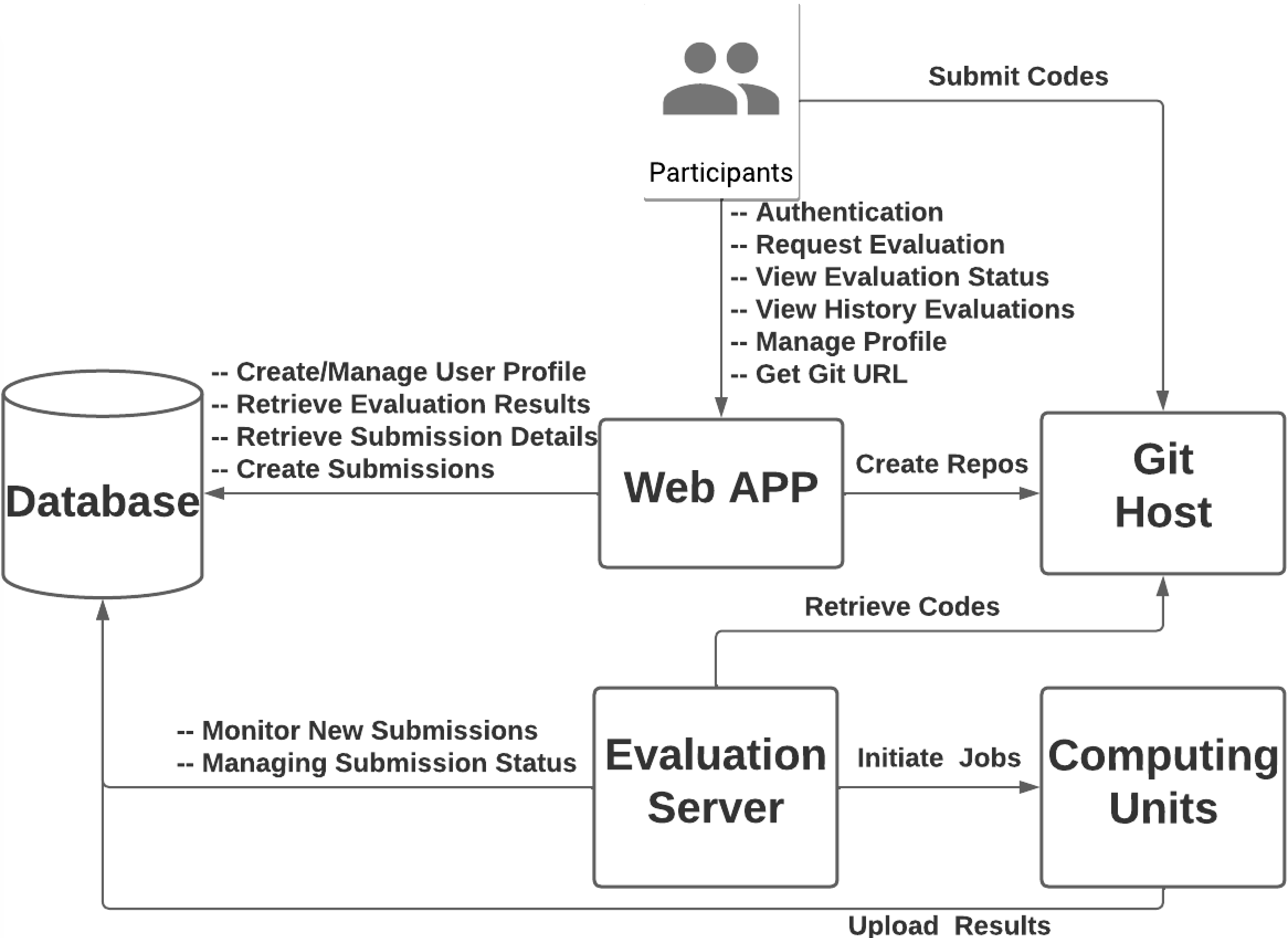}
\caption{System Architecture}
\label{fig:archi}
\end{figure}

Our competition system is designed to run as an online submission and evaluation system.
The system is responsible for managing the user profiles, submission, evaluation, and display of results.
It aims to provide a user-friendly interface for the participants to interact with the system.
The system allows participants to submit their solutions to the competition problems at any time 
before the due date, the submission will be queued for evaluation upon submission.
Participants utilise a leaderboard panel to keep track of the 
progress of the competition, and a submission history panel to view the details of each submission.

The architecture of the competition system is shown in Figure \ref{fig:archi}. 
The system consists of several main components:
\begin{itemize}
    \item a \textbf{git host} manages git repositories for participants to submit their code,
    \item a \textbf{web APP} for the users to interact with the system,
    \item a \textbf{database} stores participant profiles, submission data, and evaluation data,
    \item an \textbf{evaluation server} that monitors new submissions and initiates the evaluation jobs, and
    \item \textbf{computing units} that execute the evaluation jobs in isolated Docker containers and submit evaluation results.
\end{itemize}
The rest of this section will describe the details of each component.

\subsection{Git Host}
The git host manages git repositories that are created for each participant.
Participants submit code by pushing their solutions to their git repositories.
The evaluation server will then fetch the code from the git repository upon receiving an evaluation request. 
In practice, the repository can be stored on Github, Bitbucket, or any other git repository hosting service.

Managing submitted code using the git repositories allows efficient management of submitted codes and tracking and retrieving the history of the submissions.

\subsection{Database}
The competition database is responsible for storing user profiles, submission details, and evaluation data.
It is also the channel used by components to communicate with each other.
We use MongoDB\footnote{https://www.mongodb.com/} as the database system,
which is a NoSQL database that stores data in JSON-like documents.
The database contains the following main collections:
\begin{itemize}
    \item The \textbf{users} collection stores the user profiles, 
        including the user ID, username, email, and authentication data.
    \item The \textbf{competition} collection stores the competition details, 
        including the competition ID, competition name, competition start time, competition end time, and so on.
    \item The \textbf{subaccounts} collection stores the subaccount details for each competition a user is involved in, 
        including the subaccount ID, user ID, the git repository URL, competition ID, and competition-specific data.
    \item The \textbf{submissions} collection stores the submission details, 
        including the submission ID, subaccount ID, competition ID, submission time, evaluation status, evaluation results, and so on.

\end{itemize}

\subsection{Web APP}
The web app is the main interface for the users to interact with the system.
It contains two parts: a Restful API Node.js\footnote{https://nodejs.org} back-end and 
a single-page web app front-end built with ReactJS\footnote{https://react.dev/}.
The Restful API server is responsible for handling the requests from the web app,
 communicating with the database and communicating with the repository host, with the following functionalities:
\begin{itemize}
    \item participant registration, authentication, authorization, and profile management,
    \item submission management, 
    which creates new submission entries in the database upon receiving an evaluation 
    request from the participants,
    \item competition management, including submission details retrieval, evaluation results retrieval, and leaderboard generation.
\end{itemize}

The single-page web app front-end provides a user-friendly interface 
for participants.
It displays public information and competition guidelines to the users,
such as the competition introduction, problem descriptions, news updates, and the leaderboard to track the progress of the competition.
It also provides a user interface for the participants to register for the competition, manage their profiles,
initiate evaluations, get access to their supplied git repositories, and view their submission histories and evaluation results.

To initiate a submission, participants can find the link to their supplied git repository in the web app and push their codes to the repository.
Then by clicking the start evaluation button, the web app will create a new submission entry in the database 
and the evaluation server will be notified to initiate the evaluation process.

\subsection{Evaluation Server}
The evaluation server is responsible for monitoring new submissions and initiating the evaluation jobs when new submissions are found.
Upon a new submission request, the evaluation server goes through the following steps:
\begin{enumerate}
    \item Fetch the code from the git repository and record the commit hash of the code in the database.
    \item Initiate an evaluation job in the database.
    \item Command a computing unit to run the evaluation job.
\end{enumerate}
The evaluation server also monitors and records the status of the evaluation jobs.

\subsection{Computing Units}
The computing units are responsible for executing the evaluation jobs,
and notify the evaluation server upon the completion of the evaluation jobs.
It executes a pre-defined evaluation script and submits the evaluation results to the database.

If multiple computation units are available, 
the evaluation server can distribute the evaluation jobs to the computing units 
utilising workload manager tools like Slurm~\cite{slurm}.
In this case, a shared file system across the evaluation server and 
computing units is required to store the evaluation jobs and results.

\subsection{Docker as a Sandbox}
To ensure the security of the system and construct an isolated environment for the evaluation jobs,
computing units run the evaluation jobs in a Docker container.
Docker~\cite{docker} is a containerisation platform that allows us to run evaluation 
jobs in isolated environments.

The Docker container is built with a base image that contains the necessary dependencies for the evaluation jobs.
We allow the participants to specify the names of additional dependencies in their submission 
and the evaluation script will install them through Ubuntu's Advanced Packaging Tool (APT).

The Docker container is also configured to limit the resources available to the evaluation jobs,
such as CPU, memory, disk space, and internet access, to prevent the evaluation jobs from consuming excessive resources
and prevent malicious activities from the submitted code.

To ensure a successful compilation and execution, we provide a bash script which allows participants
to build the same docker container used for the evaluation jobs on their local machines. 
This greatly improved the efficiency of the participants in debugging runtime environment issues.

\section{Applications}

The competition system based on the proposed architecture
has been used for three different applications thus far.
In this section, we will introduce three systems that are based on the architecture
and discuss the differences across these systems.

\subsection{Teaching for an AI Planning Unit}
The first system is an online assignment submission 
and evaluation system for an AI planning unit.
The AI planning unit is a university course that teaches students
algorithms and techniques for solving planning and reasoning problems.
The unit includes two assignments that require students to solve
Multi-Agent Path Finding problems and Pac-Man Capture the Flag problems.

The system uses Bitbucket to manage the submitted code and 
uses a simple password-based authentication system.
It allows students to submit their assignment implementation
at any time before the due date.
Upon receiving an submission, the system evaluates the submission and displays the results on a leaderboard.
The evaluation is performed on a single computer.
As the accurate runtime measurement is less important here, 
we run multiple jobs in parallel, according to the number of available CPUs, to speed up the evaluation process.
Students are allowed to make another submission after receiving the evaluation results.

The system gives immediate feedback to the students
and encourages competition among the students using the leaderboard.


\subsection{The Grid-based Path Planning Competition}
The above system was then extended for the Grid-based Path Planning Competition (GPPC$^2$),
an annual competition that evaluates the performance of path-planning algorithms on grid-based maps. 
It is a forum for tracking and disseminating state-of-the-art progress in the area.
The competition measures progress in two distinct tracks:
\begin{itemize}
    \item Classical track, which is 8-connected pathfinding on a static grid map, and
    \item Any-angle track, where grid paths are not restricted to eight directions.
\end{itemize}
GPPC$^2$ uses \emph{Login with GitHub} as the authentication system and uses GitHub to manage the participant repositories. 
Participants competing in multiple tracks are supplied a git repository for each track.

\paragraph{Evaluation System:} Different from the AI planning unit, GPPC$^2$ requires precise performance measurements and 
the runtime of Grid-based Path Planning algorithms often varies due to computational interferences.
Thus the evaluation process is split into two stages on two computers: the precomputing stage and the benchmarking stage.
The precomputing stage allows participants to precompute any data structure that can be used to speed up the pathfinding process.
Precomputing jobs are often time-consuming and less sensitive to runtime variations, 
so we allow multiple precomputing jobs to run in parallel and assign even computing resources for each precomputing job.
The evaluation job is then passed to the benchmarking machine and queuing for benchmark evaluation once precomputation is done.
The benchmarking machine is a high-performance computer and only runs one job at a time to ensure the accuracy of the runtime measurement.

\paragraph{Leaderboard:} The leaderboard of GPPC$^2$ is customised to compare algorithms over various metrics, 
such as runtime, resource usage, path quality and so on.
It provides filters allowing comparisons on only undominated algorithms, 
optimal/suboptimal algorithms or precomputing/online algorithms.

\paragraph{Archives:} As the system utilises git repositories to track the submitted codes and records the commit hash for each version of evaluated codes, 
we can easily retrieve the historical versions of the submitted code to build a post-competition code archive for future reference.

\subsection{The League of Robot Runners Competition}

The League of Robot Runners Competition is 
a competition where participants tackle one of the most complex optimisation challenges:
the coordination of a large team of moving robots to fulfill tasks as efficiently as possible, 
while subject to computational constraints.

The competition uses the same Github-based authentication system as GPPC$^2$ and
uses Github to manage the participant repositories. 
The major differences in the League of Robot Runners Competition
are the evaluation cluster, a more informed leaderboard and the use of cloud computing.

\paragraph{Evaluation System:} The evaluation of the League of Robot Runners Competition requires
handling large amounts of simultaneous submissions and evaluating them in a timely manner.
The evaluation server is connected to a cluster of computing units managed by Slurm Workload Manager~\cite{slurm}.
The evaluation server submits the evaluation jobs to the Slurm Workload Manager,
which then schedules the jobs to the computing units.
The computing units are virtual machines running on AWS EC2 instances, 
and they will execuate a predefined evaluation script to evaluate the submitted codes, 
submit the evaluation results to the database, and back up the raw output files to S3 cloud storage\footnote{https://aws.amazon.com/s3/}.
Using the AWS ParallelCluster service\footnote{https://aws.amazon.com/hpc/parallelcluster/}, computing units are dynamically allocated upon the demand of jobs,
we allow the system scaling to 12 computing units during the peak submission period.

\paragraph{Leaderboard:} The leaderboard used in the competition classifies results into three distinct categories,
the overall best category, the line honors category, and the fast move category,
and each with a different scoring function to rank participants.
It also provides visualisations of the history of the competition, 
allowing participants to compare their performance with others over time.
Additionally, we provide an all-submissions tab to allow participants to monitor
who is submitting and how much progress they have made, which enhances the competition atmosphere.

\subsection{Applicability to Other Competitions}

The proposed competition system should apply to other competitions as long as the 
workflow of the competition includes submitting, evaluating, and leaderboard generation.
However, the system may require modifications to adapt to the specific requirements of the competition,
for example, the evaluation process, evaluation results processing, and the leaderboard generation
could be different for different competitions.
The differences in the evaluation output data lead to most of the changes when moving from GPPC$^2$ 
to the League of Robot Runners Competition,
as the processing of these data is required across the system.

\section{Issues and Challenges}

When designing and running these competitions, we encountered several issues and challenges.
This section will discuss some of the issues and challenges we faced and how we addressed them.

\paragraph{Multiple Tracks:} Many competitions have multiple tracks to evaluate different aspects of the problem.
In the GPPC$^2$ competition, participants can compete in both the classical track and the any-angle track.
We treat each track as a separate competition entry in the Competition collection of the database. 
After login, participants are allowed to join and create subaccounts for 
each track, with each subaccount tagged with the corresponding competition ID.
Each subaccount will be supplied a separate git repository for each track.
Two leaderboards display the results of each track separately.

In contrast, in the League of Robot Runners Competition
each round of the competition runs as a separate competition entry.
The different categories on the leaderboard look like several tracks, but they just 
sort participant teams based on different criteria using the same submission data.

\paragraph{Periodically Competitions:} Most competitions are held periodically. 
One can achieve this by creating a new competition entry in the Competition collection of the database
or independently host the system for each competition. 
In the teaching unit, we simply archive and reset the database and git repositories after each semester.
The GPPC$^2$ is a rolling competition without problem changes on existing tracks, thus we just create submission archives and results summaries across submissions from different years.
The League of Robot Runners Competition is held annually, and we plan to create new competition entries for future years.

\paragraph{Export Submissions and Solutions: }The competition organizers may require the export of solutions and submissions to verify the results
and build a post-competition archive. 
Retrieving submissions is easy as each submission records the commit hash of the evaluated code, 
together with the git repository that hosts the submitted codes, so one can easily check out each version of the submitted code.
If the competition requires recording the solutions produced by the submitted code, 
we could simply write these solutions to the database (small output), a hard drive, or in the cloud storage, 
and organise them by the submission ID.

\paragraph{Debugging:}Debugging participant submissions is another challenge, as this needs to trade-off transparency with the feasibility of debugging.
In GPPC$^2$, we provide debug instances that are similar to the evaluation instances,
and each submission will be evaluated on these debug instances first, followed by the evaluation instances.
All the output and logs on the debug instances are accessible to the participants, 
but all output and logs on the evaluation instances are hidden from participants.
In the League of Robot Runners Competition,
we provide a large number of example instances for participants to debug their implementation offline,
and hide all evaluation logs and output from the participants.
Only the server operation logs on the submission are accessible to the participants.
If participants cannot solve their problem with the provided debug instances or offline example instances,
they can seek help from the competition organisers. The organisers can view the evaluation logs and output to provide debugging support. 

\paragraph{Multiple Entries:} In GPPC$^2$, some participants would like to submit multiple algorithms to be displayed on the leaderboard.
We allow participants to create multiple subaccounts for each track (competition) 
and each subaccount will have a place on the leaderboard and with independent git repositories.
But, to prevent malicious activities from flooding the leaderboard, by default, 
each participant only has one subaccount for each track. 
They can then request permission to create more subaccounts for the same track, 
with a limit on the maximum number of subaccounts.

\paragraph{Compute Limitations:} Computation resources are always limited,
and the evaluation process may take a long time to complete even on broken implementations. 
In GPPC$^2$ and the League of Robot Runners, 
the evaluation server will kill an evaluation job if it exceeds a certain time limit.
CPU limits, Memory Limits, and Disk I/O limits are all applied through the docker container configurations.

\paragraph{Ranking Solutions:} The competition may require ranking solutions based on multiple criteria.
In GPPC$^2$, we do not rank solutions directly. Instead, we list all the metrics we collected on the leaderboard,
and anyone can utilise the sorting function, and a variety of filters to rank the solutions based on their preferences.
The aim is to not diminish the importance of any metric and let investigators decide 
which metric is more important according to the use case and the challenges they are facing.
The League of Robot Runners Competition utilises multiple scoring functions to rank the solutions in different categories.
In this way, we produce the winners and the competition covers a wider range of optimisation challenges.

\paragraph{Auditing and Rules:} The competition system tries to prevent malicious activities and cheating by isolating the evaluation environment and overwriting unmodifiable files in the submitted code.
However, these measures are limited, one can always find a way to cheat the system; 
the cost of prohibiting cheating is high.
Thus, we rely on an auditing process to detect cheating activities 
and setup rules to exclude participants who attempt to interfere with, or otherwise attempt to hijack or misappropriate, 
any part of the evaluation system/server functionalities.
At the end of the competition, we manually review the submitted codes of the 
top participants to ensure the fairness of the competition.

\section{Conclusion and Future Work}

Managing competitions is not easy, with many challenges to overcome and efforts to put in. 
In this paper, 
we present an online competition system that allows participants to submit their solutions 
to the competition problems at any time before the due date.
The competition system utilises Git repositories to manage submitted codes, 
a Web APP for participant interaction,
Slurm to manage computing units, and Docker to create an isolated evaluation environment.
We then discuss the issues and challenges we faced while operating and developing the competition system.

Currently, the system used for the three applications is maintained separately,
which increases the workload for maintaining these systems and the difficulty of adding new features.
Thus one target of future work is to better modularise and generalise the implementation of the system
so that the common components can be maintained in one place and shared among the applications. 
The system will then be open-sourced as a community resource to help researchers in related fields host competitions.

\bibliography{aaai24}

\end{document}